\documentclass[journal]{IEEEtran}
\usepackage[noadjust]{cite}

\ifCLASSINFOpdf
  \usepackage[pdftex]{graphicx}
\else
\fi
\usepackage[cmex10]{amsmath}
\usepackage{url}

\usepackage{comment}

\usepackage{psfrag}
\usepackage{pstool}

\begin{document}

\onecolumn

This work has been submitted to the IEEE for possible publication. Copyright may be transferred without notice, after which this version may no longer be accessible

\twocolumn
\newpage

%
\title{Obstacle Avoidance Strategy using Onboard Stereo Vision on a Flapping Wing MAV   }
%
%
%

\author{S.~Tijmons, G.C.H.E. de Croon,~\IEEEmembership{Member,~IEEE}, B.D.W. Remes, C. De Wagter, M. Mulder
\thanks{All authors are with the Control and Simulation Section, Faculty of Aerospace Engineering, Delft University of Technology, The Netherlands \mbox{(e-mail: s.tijmons@tudelft.nl, g.c.h.e.decroon@tudelft.nl)}}
}

%
%

\markboth{IEEE TRANSACTIONS ON ROBOTICS}%
{Shell \MakeLowercase{\textit{et al.}}: Bare Demo of IEEEtran.cls for Journals}
%



\maketitle

\begin{abstract}
The development of autonomous lightweight MAVs, capable of navigating in unknown indoor environments, is one of the major challenges in robotics. The complexity of this challenge comes from constraints on weight and power consumption of onboard sensing and processing devices. In this paper we propose the ``Droplet" strategy, an avoidance strategy based on stereo vision inputs that outperforms reactive avoidance strategies by allowing constant speed maneuvers while being computationally extremely efficient, and which does not need to store previous images or maps. The strategy deals with nonholonomic motion constraints of most fixed and flapping wing platforms, and with the limited field-of-view of stereo camera systems. It guarantees obstacle-free flight in the absence of sensor and motor noise. We first analyze the strategy in simulation, and then show its robustness in real-world conditions by implementing it on a 20-gram flapping wing MAV.

\end{abstract}

\begin{IEEEkeywords}
Collision avoidance, aerial robotics, stereo \newline vision, micro robots
\end{IEEEkeywords}


%
\IEEEpeerreviewmaketitle

\section{Introduction}
%
%
%
%

\IEEEPARstart{A}{utonomous} flight of Micro Air Vehicles (MAVs) is an active area of research in robotics. Because of its wide scale of potential applications it is gaining a growing amount of attention from governments and industry. Especially for outdoor applications, such as surveillance, monitoring, aerial photography and mapping, many commercial MAV systems are currently available. For indoor applications, however, this is not so much the case as these systems require more advanced methods for localization and navigation. This forms a challenge as more onboard sensors are required while indoor applications often require small sizes for the platforms. For extremely lightweight MAVs under 50~g many solutions for autonomous navigation from the literature are therefore not applicable. Active sensors such as laser range finders \cite{bachrach2011range,grzonka2012fully,nieuwenhuisen2015autonomous} and RGB-D cameras \cite{huang2011visual} are typically heavier than the 50~g platforms. Cameras, which are passive, are commonly used in combination with Simultaneous Localization and Mapping (SLAM) methods \cite{fraundorfer2012vision,mur2015orb,engel2014lsd}, dense reconstruction methods \cite{pizzoli2014remode} or visual odometry methods \cite{forster2014svo} which provide information needed for obstacle avoidance, localization and navigation. A downside of these methods is their high demand for processing power and memory.


Obstacle avoidance and also other aspects of indoor navigation have nonetheless been demonstrated on several lightweight platforms. In most of these studies, with platforms ranging from 10~g to 30~g, onboard sensing is realised by using optical flow sensors \cite{moore2014autonomous,hyslop2010autonomous,ross2013learning,beyeler2009vision,zufferey2010optic,zufferey2010autonomous,conroy2009implementation}. These sensors can be very light and fast, but at the cost of providing low resolution. As a consequence, flow inputs generated by small obstacles are filtered out \cite{hyslop2010autonomous}. Optical flow also has the limiting properties that no flow information is available around the focus of expansion (direction of motion) \cite{beyeler2009vision}, and that the flow estimates only provide relative speed information. As a consequence mainly reactive methods have been applied in these studies, that use the differences between measurements from multiple sensors to balance the distances to surrounding objects. This provides effective methods for specific guidance and avoidance tasks, but does not guarantee collision-free flight.


Stereo vision is considered to be a more robust method for the purpose of obstacle avoidance. The main advantages are that objects can also be detected in the direction of motion and that it provides absolute distance estimates to these objects. These advantages were demonstrated on a relatively heavy platform equipped with optical flow and stereo vision systems \cite{hrabar2005combined}. Onboard stereo vision has also been demonstrated on a fixed-wing vehicle of over 500~g flying at 9~m/s \cite{barry2014pushbroom}. In flight tests small obstacles right ahead of the vehicle are detected at a range of 5~m and at a frame rate of 120~Hz. This approach shows that stereo vision can be used to combine short-term path planning with reactive avoidance control. 


The \textbf{main contributions} of this study are the introduction of a computationally efficient avoidance strategy, its validation by simulation experiments and its implementation and validation on an extremely light flapping wing MAV. The avoidance strategy ensures obstacle avoidance, even in complex and closed environments, based on information from a stereo vision system. The strategy is specifically designed for narrow and extremely lightweight systems, flying in narrow and cluttered environments that are restricted to, or prefer to maintain, a minimum forward velocity while having a limited turn rate (nonholonomic constraint). The strategy explicitly takes into account the limited field-of-view of the cameras. The method does not require to create a map or to store camera images or disparity maps. This combination of properties results in an efficient and effective method for obstacle avoidance that is suitable for implementation on tiny, embedded systems. This is validated by computer simulations and real flight experiments.



A preliminary version of the avoidance strategy was tested on a flapping wing MAV that relied on off-board processing \cite{tijmons2013stereo}. The flapping wing MAV in the current study, which includes onboard processing, previously demonstrated a standard method for reactive avoidance \cite{de2014autonomous}. The current study presents a number of important improvements over our previous work. First, the ``Droplet" strategy is introduced, which incorporates a new set of decision rules that improves its robustness, by taking into account the limitations of the onboard stereo vision system. In addition, the theory behind the strategy is described in detail, a theoretical proof of guarantee is presented and its computational efficiency is compared to related approaches. A different stereo vision algorithm running on board of the vehicle is presented that improves robustness, reliability and efficiency. The proposed avoidance strategy is evaluated through extensive computer simulations to show the effectiveness of the method in order to compare it with other reactive methods that have comparable computational complexity, and to analyze its performance in combination with the actual obstacle detection sensor. Furthermore, the avoidance strategy is demonstrated experimentally through test flights with the DelFly Explorer, a 20~g flapping wing MAV with onboard stereo vision processing. This is the first study showing obstacle avoidance based on an onboard stereo vision system with a real platform with such a low weight.


The article is organized as follows. Section~\ref{related} discusses related work on other flapping wing MAVs and different obstacle avoidance methods. In Section~\ref{system} the DelFly Explorer and its onboard stereo vision system are described. In Section~\ref{strategy} the avoidance strategy is explained. The avoidance strategy is compared to other methods and analyzed through simulations in Section~\ref{simulations}. Flight experiments with the real platform and vision system are evaluated in Section~\ref{experiments}. Finally, conclusions are drawn in Section~\ref{conclusions}. The Appendix contains a section describing how the avoidance maneuver that is presented in this study can be extended to 3D.

\section{Related work} \label{related}

The number of studies on autonomous capabilities of flapping wing MAVs is fairly limited, as much research focuses on the design of such vehicles. Their lightweight designs limit the possibilities to use onboard sensors, and many studies demonstrate guidance and control capabilities using ground-based tracking systems \cite{lin2010altitude,hsiao2012using,BAEK2010,julian2013cooperative,DECROON2009,MA2013}. However, a number of studies also demonstrate control, guidance and navigation tasks on flapping wing vehicles. IMU-based attitude stabilization has been demonstrated on a 19~g \cite{KEENNON2012} platform and attitude control on a 0.1~g platform \cite{helbling2014pitch,fuller2014using}. The latter also demonstrated height control using an optical flow sensor \cite{DUHAMEL2012}. Guidance tasks have been realized on 16~g platforms, such as target-seeking, using an onboard Wii-mote infrared camera \cite{BAEK2011b}, and line following, using an onboard camera and an off-board processing unit \cite{DECROON2009}. Autonomous navigation-related tasks demonstrated so far are vision-based obstacle avoidance indoors (but using off-board processing) with a 16~g platform \cite{DECROON2012C} and GPS/IMU-based loitering outdoors with a 312~g platform \cite{roberts2014autonomus}. To our knowledge, autonomous obstacle avoidance has not been demonstrated on a flapping wing MAV.

As mentioned in the introduction, many studies perform autonomous navigation on other platform types. On lightweight platforms a common approach is to use reactive control based on optical flow sensors \cite{moore2014autonomous,hyslop2010autonomous,ross2013learning,beyeler2009vision,zufferey2010optic,zufferey2010autonomous,conroy2009implementation}. Several studies on reactive methods mention the possibility of collisions with obstacles outside the field-of-view of the sensors \cite{ross2013learning},\cite{bipin2015autonomous}. On heavier platforms either heavier active sensors such as laser rangefinders are used \cite{bachrach2011range,grzonka2012fully,nieuwenhuisen2015autonomous,huang2011visual}, or computationally demanding vision methods are applied, such as SLAM \cite{engel2014lsd,mur2015orb}, dense reconstruction \cite{pizzoli2014remode} or visual odometry \cite{forster2014svo}, using monocular or stereo cameras. These methods provide relative or absolute ranges to points in the environment which form the basis for localization and obstacle detection. This information enables path planning, which can be more robust than reactive methods, but at the cost of more computational load. A closed-loop rapidly exploring random tree (RRT) approach has been demonstrated in combination with stereo vision and SLAM on a 1~kg quadcopter \cite{matthies2014stereo}. This approach uses an efficient algorithm to check for collisions in disparity space while generating trajectories to candidate waypoints. Another study also focuses on using an efficient representation of obstacle locations and possible vehicle states by using an octree-based state lattice \cite{xu2015real}. This keeps memory consumption low and makes their $A^{\ast}$ graph search for finding an optimal trajectory more efficient. The method is demonstrated onboard a quadrotor that is equipped with a stereo vision system for producing disparity maps.

This study uses motion primitives, which is an efficient method for generating candidate trajectories as the method relies on a set of pre-computed control input sequences and trajectories which are checked for collisions. This has also been demonstrated on a real quadcopter equipped with a LIDAR sensor using a graph search algorithm \cite{macallister2013path}. A similar approach for path planning has been proposed for very small MAVs \cite{pivtoraiko2013incremental} but this study does not show experiments with a real perception sensor. Another study that also focuses on using motion primitives on very small platforms makes use of a receding horizon control-based motion planner, but again no real perception sensor was used in the experiments \cite{paranjape2015motion}. Their approach uses two types of motion primitives: steady turns for waypoint connection and transient maneuvers for instantaneous heading reversal. This combination of possible actions ensures the availability of an escape route, which is a shortcoming of reactive methods. Instead of storing a map with all detected obstacles, it has been proposed to use a local map of surrounding obstacles in combination with a fading memory, which remembers obstacle locations for a while, and deletes them later on \cite{dey2014vision}. This restricts memory requirements while allowing a planner to take into account obstacles outside the field-of-view.

In this paper we propose an avoidance strategy that guarantees obstacle avoidance and the availability of an escape route for a vehicle with nonholonomic constraints, without the need of storing disparity information or a map with obstacle locations. The strategy is particularly beneficial in terms of robustness for systems that rely on small embedded perception sensors, such as the stereo vision system used in this study, as such sensors have various limitations: low resolution, sensitivity to motion (image blur), limited field-of-view, limited range. Furthermore the proposed avoidance strategy relies on an efficient algorithm that checks for collisions in disparity space and uses a small set of rules within a finite-state machine to make turn decisions.

\section{System Description}\label{system}

\begin{figure}
\centering
  \includegraphics[width=0.485\textwidth]{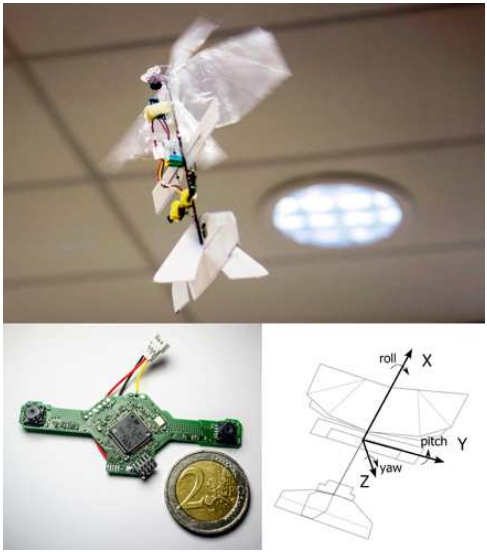}
\caption{The DelFly Explorer platform. \textbf{Top}: The DelFly Explorer in slow forward flight condition. \textbf{Bottom-left}: Closeup image of the custom-made stereo vision camera, baseline 6 cm. \textbf{Bottom-right}: body-axes definition. In slow forward flight the body X-axis points upward and the Z-axis points forward.}
\label{figure:DelFlyExplorer}       
\end{figure}

The avoidance strategy proposed here is tested on the DelFly Explorer, the Flapping Wing MAV shown in Fig.~\ref{figure:DelFlyExplorer}. This platform has a wingspan of 28~cm and a weight of 20~g. The Explorer version of the DelFly has a few important differences compared to its design used in previous studies (\cite{DECROON2012C,tijmons2013stereo}). First, it has a cumston-made 1.0~g autopilot board which includes all electronics required for flight control: an electronic speed controller for its brushless motor, a transceiver for two-way communication with a remote station, and an ATmega328P 20~MHz microcontroller for onboard processing. The microcontroller has access to an MPU9150 9-axis IMU (gyro-,~accelero-,~magnetometers) and a BMP180 pressure sensor, which are used for attitude and altitude control. 

A second difference is the addition of actuated aileron surfaces which provide more turn rate authority by creating an aerodynamic moment on the body X-axis (see Fig.~\ref{figure:DelFlyExplorer}). At high pitch angles the ailerons are therefore more effective for horizontal heading control than the original tail rudder which acts on the vehicle's Z-axis.


The DelFly Explorer is further equipped with a custom-made stereo vision camera system of 4~g which is used for obstacle detection in this study. The camera system runs a stereo vision algorithm to obtain a disparity (inverse depth) map, based on which it is decided whether the vehicle needs to perform an avoidance maneuver or not. This information is communicated to the autopilot which controls the avoidance maneuvers. 

\subsection{DelFly Explorer flight characteristics} \label{delfly}

The DelFly Explorer shares similarities with fixed-wing aircraft but it has two important differences: the wings generate not only lift but also thrust, and the location of the center of gravity is further aft, close to the trailing edges of the wings. This enables the vehicle to fly passively stable at low forward speeds and at high pitch angles. This is illustrated by the top image in Fig.~\ref{figure:DelFlyExplorer}. The forward flight velocity is typically in the range of 0.5-1.0~m/s and the pitch angle in the range of 70-80$^{\circ}$. Forward velocity and pitch angle are controlled by the elevator and the motor speed. When the vehicle hovers or flies backward it is not passively stable.

The vehicle possesses a nonholonomic constraint in that it has no authority over its lateral velocity along the Y-axis. In the horizontal plane it can only be guided by controlling the heading angle, for which the ailerons are used. The lateral velocity is damped by the surfaces of the wings and the vertical tail. Lateral drift is therefore determined by the velocity of the air. The vertical velocity is mainly controlled by the motor speed which determines the flapping frequency.

The lift forces produced by the wing and tail surfaces at high pitch angles, even at such a low forward velocity, significantly improve flight efficiency. Flight times of over 9 minutes have been recorded in this study at a forward speed of 0.6~m/s and using a Li-Po battery of 180~mAh. When hovering, the flight time reduces to around $3$ minutes. This property forms an advantage of a flapping wing design over the more conventional quad rotor design which is commonly applied for indoor tasks. For example the 19~g Crazyflie Nano Quadcopter\footnote{\url{https://www.bitcraze.io/}} can fly for up to 7 minutes on a 170~mAh Li-Po battery. When a 6~g camera system is added, the flight time reduces to 3.5 minutes \cite{dunkleyvisual}. To benefit from the higher flight efficiency, the avoidance strategy proposed here guides the DelFly such that it can maintain its forward velocity while avoiding collisions. At the same time, the vehicle also benefits from the passive stability characteristics when flying forward.

\subsection{Stereo vision system} \label{vision}

A common approach for obstacle avoidance and navigation tasks on small platforms is the use of optical flow systems \cite{moore2014autonomous,hyslop2010autonomous,ross2013learning,beyeler2009vision,zufferey2010optic,zufferey2010autonomous,conroy2009implementation}. However, for the task of obstacle avoidance, stereo vision has several advantages over optical flow. First, stereo vision provides true scale (instead of relative scale) estimates of distances to obstacles. Second, distance estimates are obtained using images from the same point in time, instead from different points in time. For stereo vision, this means that image points between two frames can only shift in one direction, while for optical flow, image points can shift in any direction. The number of image points appropriate for matching is therefore much higher for the case of stereo vision, and the matching process is also more efficient. Third, optical flow is small close to the Focus of Expansion, the image region that is in line with the direction of motion of the camera. Hence, small flow inaccuracies can have a large deteriorating effect on obstacle detection in the crucial flight direction. Stereo vision also provides reliable distance estimates for image features in this image region.

The custom-made vision system onboard the Explorer is shown in Fig.~\ref{figure:DelFlyExplorer}. It features two 30~Hz CMOS (TCM8230MD) cameras with 640$\times$480 pixels resolution and a field-of-view of 58$^{\circ}\times45^{\circ}$. An STM32F405 microprocessor performs the stereo image processing. Memory restrictions (192~kB~RAM) and processing restrictions (168~MHz) constrain the stereo vision algorithm to only use a sub-resolution of 128$\times$96 pixels for calculating disparity maps. Disparity maps contain disparity (inverse distance) values per image pixel and thereby indicate the presence and location of obstacles.

To ensure computational efficiency, a sparse stereo vision method is implemented. The method is based on standard Sum of Absolute Differences (SAD) window matching \cite{SCHARSTEIN2002}. For efficiency reasons, window matching is only tried at image locations where the horizontal image gradient (determined by horizontal differential convolution over single lines) contains a local peak and if this gradient exceeds a predefined threshold. For robustness reasons the output from the window matching computation is evaluated using a peak ratio test (see \cite{hu2010evaluation} for a comparison with other confidence metrics). Only if the ratio between the cost of the best match and the second-best match (excluding direct neighbors of the best match) exceeds a predefined threshold, the match is regarded reliable. Sub-disparity estimates are then calculated using parabola fitting using the three matching costs around the best match. 

The method runs at a frame rate of 15-30~Hz. This rate varies with the amount of texture observed in the image. Note that at least some form of texture is required to detect an obstacle. In this study sufficient texture is added to the obstacles as our focus is on the proposed avoidance strategy. 

\begin{figure}[!t]
\centering
  \includegraphics[width=0.49\textwidth]{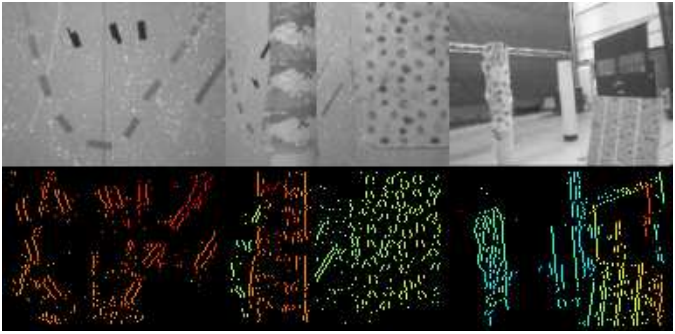}
\caption{Example disparity images obtained from the stereo vision system using the sparse block matching method as used in this study. Black pixels are shown when no matches are found for these pixels. Red pixels represent high disparity values (small distances) and blue pixels represent low disparity values (large distances). Yellow and green represent medium values for disparity. The figure is best viewed in digital format.}
\label{disp_maps}       
\end{figure}

The stereo vision algorithm is different from the method that was implemented in previous research on reactive obstacle avoidance \cite{de2014autonomous}. The algorithm in the previous study is intended to deliver dense disparity maps in an efficient way, and is specifically tuned to provide sensible disparity estimates in image regions that lack texture (e.g., smooth walls). However, the resulting disparity maps contain a lot of bad matches due to the assumption of fronto-parallel planes and also contain a relatively high degree of noise. The method produces a lot of noise specifically in image regions that contain dense texture, which is undesirable. Furthermore the quality of the estimated disparity values cannot be monitored and can vary considerably within a single image. The algorithm implemented in the current study produces only sparse disparity maps but solves the main issues of the previous method: it returns disparity estimates only for points with relatively high certainty, the certainty is higher in texture-rich image regions and the number of bad matches is significantly reduced. As a result, sparse disparity maps are produced that contain disparity values with a relatively high certainty and accuracy, while the number of computed disparity points can be used as a measure of reliability. Furthermore, the average frame rate is also higher compared to the previous method ($>$15~Hz compared to $\sim$11~Hz).



Fig.~\ref{disp_maps} shows examples of the sparse disparity maps computed by the stereo vision system. These examples illustrate how much depth information is obtained and how this relates to the quantity of image texture. The examples also serve as a way to show that the quality of depth information is more important than its quantity. The most right image serves as a good example. The left part of this image contains a sparse amount of information. Yet it can be assumed that no near obstacles are present on this side of the image, while on the right side of the image near obstacles are present.

\begin{figure}
\centering
  \includegraphics[width=0.485\textwidth]{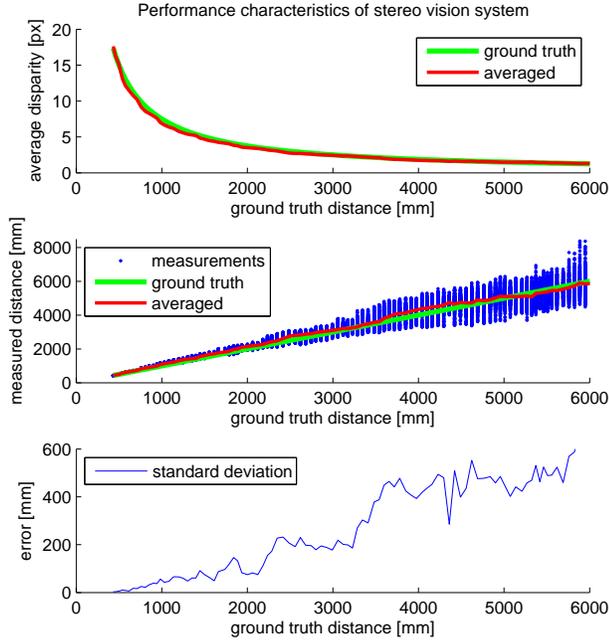}
\caption{Performance characteristics of the camera system shown in Fig.~\ref{figure:DelFlyExplorer}. Estimates are based on 150-300 stereo matches per frame. \textbf{Top}: average estimated disparity based on all matches per frame. \textbf{Middle}: Spread of estimated distances from individual stereo matches, as well as averaged estimated distance. \textbf{Bottom}: Standard deviation of the distance error.  }
\label{vision_characs}       
\end{figure}

Characteristics of the camera system performance are shown in Fig. \ref{vision_characs}. These results give an insight in the accuracy with which the system can detect distances to obstacles, and how this accuracy declines with increasing distance. Note that at a distance of 3 meters, the standard deviation of the estimated distance is around 200~mm. This observation is important as this distance corresponds to the maximum that needs to be observed in the experimental flight tests.



\section{Avoidance Strategy} \label{strategy}

\begin{figure}[ht]
  
  \includegraphics[width=0.48\textwidth]{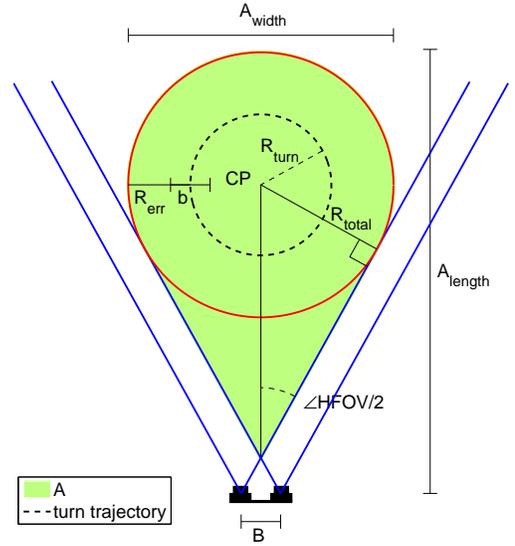}
\caption{Top-view of the Droplet avoidance area showing its contours and parameters that define its shape. The filled (green) area defines the region within the field-of-view of the cameras (indicated by the black symbols at the bottom) that needs to stay clear of obstacles. The shape of this area is defined such that a vehicle is able to fly circles within this region (as indicated by the dashed line). The Droplet region is defined in the camera reference frame. }
\label{droplet_params}       
\end{figure}

\subsection{Avoidance maneuver} \label{avoidance_maneuver}
The starting point of the proposed avoidance strategy is that, when avoiding obstacles, the vehicle performs steady turns with a constant flight speed and a constant turn rate. By maintaining its forward speed the vehicle benefits from a higher flight efficiency, better stability and sufficient response to aileron control inputs, as explained in Section~\ref{delfly}. This all means that the vehicle will perform avoidance maneuvers with a constant turn radius $R_{turn}$ given by:

\begin{equation}
R_{turn} = V/\dot{\psi} 
\label{turn_radius}
\end{equation}

Both forward speed $V$ and heading turn rate $\dot{\psi}$ primarily depend on vehicle dynamics. In Section~\ref{parameters} we will show, however, that also the update rate of the stereo vision system affects the range of possible velocities and turn rates that can be set. Fixed values are assumed for the forward speed, turn rate and turn radius. To avoid collisions, the vehicle needs a sufficiently large circular turn space when performing maneuvers, with radius:


\begin{equation} 
\begin{aligned}
R_{total} = R_{turn} + b/2 + R_{marg}
\end{aligned}
\label{radius}
\end{equation}

Here $b$ is the vehicle wingspan and $R_{marg}$ is an error margin to account for deviations from the avoidance trajectory and inaccuracies in measured distances to detected obstacles. 




The novelty of the method presented here is that it continuously checks if there is such a turn space ahead of the vehicle that is free of obstacles. This is illustrated in Fig.~\ref{droplet_params} by a top-view schematic. The turn space with radius $R_{total}$ is indicated by the red circle with center point $CP$. The position of $CP$ is at some distance ahead of the camera such that the turn space fits within the combined field-of-view $HFOV$ of the cameras, which is defined as the overlapping part of the two blue cones. The green area $A$ then defines the minimum area that needs to be free of obstacles in order to guarantee a safe avoidance maneuver. Since this area has the shape of a Droplet, we call our method the ``Droplet" strategy. The avoidance maneuver will be initiated as soon as an obstacle is detected inside the Droplet region. 

In contrast to many other avoidance maneuvers found in the literature, our maneuver does not only guarantee collision avoidance up to the end of some proposed path. In fact, our method proposes an infinite path as it makes sure that the vehicle can fly the circular turn trajectory indicated in Fig.~\ref{droplet_params}.

The need for finding a free space ahead of the vehicle, stems from the fact that the cameras have a limited field-of-view. This issue can be tackled by using multiple camera systems pointing in all directions or by using wide-angle/panoramic lenses. Instead of adding extra payload weight by using one of these approaches, the proposed method only requires the execution of a few additional control rules.

The distance $CP_{dist}$ between the camera and center point $CP$ is given by:

\begin{equation}
CP_{dist} = \frac{R_{total}}{\sin(HFOV/2)}+\frac{B/2}{\tan(HFOV/2)}
\end{equation}

$B$ is the baseline of the camera system, which is 60~mm. The size of the Droplet area $A$ is important as it defines the minimum size of spaces that can be accessed by the vehicle. The outer dimensions of this space are given by:

\begin{equation}
\begin{aligned} 
A_{width} = 2R_{total} \\ 
&\quad \\
\end{aligned}
\label{eqn:width}
\end{equation}

\begin{equation}
\begin{aligned} 
 &\quad \qquad \qquad  A_{length} = CP_{dist}+R_{total}\\ 
&\quad =R_{total}\left (1+\frac{1}{\sin(HFOV/2)}\right )+\frac{B/2}{\tan(HFOV/2)} \\
&\quad \\
\end{aligned}
\label{eqn:length}
\end{equation}

\begin{figure}
  \includegraphics[trim={0.5cm 0 0cm 0}, clip,width=0.5\textwidth]{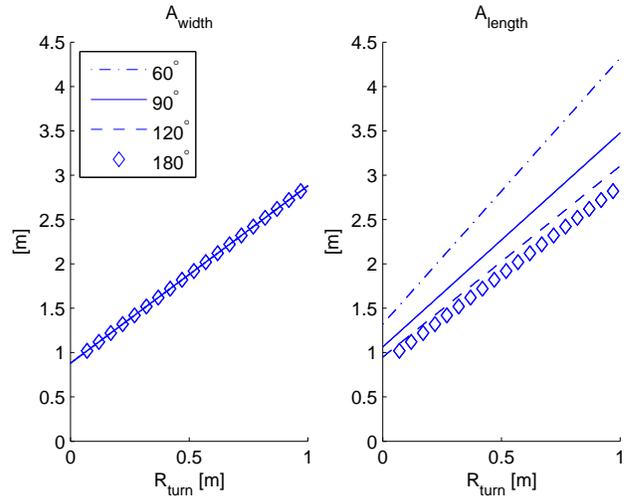}
\caption{Dimensions of the Droplet avoidance area. These dimensions are dependent on the turn radius ($R_{turn}$) of the vehicle and the horizontal field-of-view of the camera ($HFOV$). The width $A_{width}$ is shown at the left, which is independent of the size of $HFOV$. The length $A_{length}$ is shown at the right.  }
\label{droplet_length_area}       
\end{figure}

The relations from Equations~\ref{eqn:width} and \ref{eqn:length} are visualized in Fig.~\ref{droplet_length_area}. The $HFOV$ values are typical for cameras with normal to wide-angle lenses. For the turn radius a range of values is shown that fits to the characteristics of the vehicle from this study. As mentioned in Section~\ref{delfly} the forward speed is typically 0.5-1.0~m/s, while lower speeds are possible but not desirable for several reasons. At turn rates of 1-2~rad/s the value of $R_{turn}$ would be in the range of 0.25-1.0~m which proved to be realistic numbers in real test flights. The wingspan $b$ of the DelFly is 28~cm and the value of the error margin $R_{marg}$ is set at 30~cm.

\subsection{Obstacle detection rules} \label{detection_rules}

The task of the stereo vision system is to detect whether there is any obstacle present within the Droplet area. Based on the shape of this region, a threshold disparity value can be calculated for each camera pixel such that the Droplet shape is defined in a reference disparity map. As the Droplet shape is static, this reference disparity map can be precomputed, making the obstacle detection process computationally inexpensive. By comparing each new observed disparity map with the reference disparity map, the number of pixels can be counted that exceeds the reference value. If this number is higher than a threshold $\tau_{d>ref}$ (=7~px in our experiments) it is assumed that an obstacle has entered the Droplet area. 


This approach forms the bare implementation of the Droplet method. This implementation would be sufficient to perform obstacle avoidance in an ideal case (large, well-textured obstacles, perfect sensing), but leads to detection failures in many real world situations. To improve robustness in those cases, the bare detection method is extended in two ways.

First, the sparsity of the disparity maps is used as a quantitative measure of image texture. The stereo vision algorithm produces only disparity values for pixels at image locations with high intensity gradients. The number of pixels having a disparity value is therefore counted in the left and right image halves individually. If either of the two sides contains less disparity values than a minimum threshold $\tau_{\#d_{\textrm{min}}}$(=50~px), the observation is regarded as if an obstacle would be detected. This rule deals with situations where a texture-poor surface (such as a white wall) is approached. The disparity images can be split in several numbers of sub-images and in multiple ways, but splitting in two halves turned out to be very effective according to experiments.


The second adjustment deals with situations where objects are hard to observe, for instance if only the edge of an obstacle is visible. This occurs when objects do not contain sufficient texture or if their appearance resembles the appearance of the background. If a certain obstacle is only partly observed and only in some of the frames, the probability of detecting that obstacle can be improved by using disparity information from a series of subsequent frames. This is realised by taking into account the distances with which individual points have moved into the Droplet area. By the assumption of constant velocity, individual estimates of these distances can be integrated over subsequent frames. By discretization of these estimates,  an array of ``obstacle-detection counters" is then obtained. Each obstacle-detection counter forms a measure for the probability that an obstacle is present at a certain distance and is used further on to define at what point an avoidance maneuver needs to be initiated. This method is based on the idea that obstacle detections from noise lead to predictions with a large spread in estimated distances, while obstacle detections belonging to a single object lead to predictions that are concentrated around a certain distance.

\begin{figure}[t]
\centering
  \includegraphics[trim={0.6cm 0 1.2cm 0}, clip, width=0.45\textwidth]{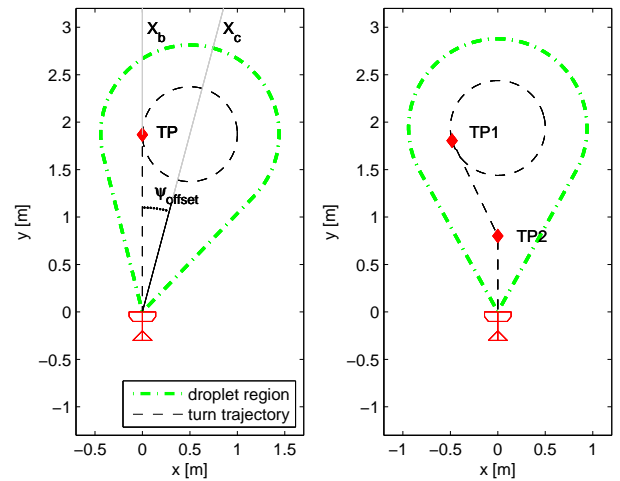}
\caption{\textbf{Left:} Visualisation showing how the Droplet strategy is implemented on the vehicle. The dashed (black) line shows the trajectory of the vehicle once the avoidance maneuver is initiated. The camera offset angle $\psi_{offset}$ shows how the camera is mounted such that the Droplet region (indicated by the dash-dotted/green line) encloses the avoidance trajectory. The turn point $TP$ indicates the point where the vehicle starts turning if the avoidance sequence is initiated. Note that the trajectory is defined in the body reference frame of the vehicle. \textbf{Right:} Alternative implementation of the Droplet strategy where the camera system is aligned with the body axis of the vehicle. This requires an extra turn point ($TP2$) where the vehicle steers towards $TP1$ in order to follow the circular trajectory. }
\label{droplet_maneuver}       
\end{figure}

\subsection{Implementation}

Based on the safety region as defined by the Droplet region in Fig.~\ref{droplet_params}, simple control rules are formulated to perform robust obstacle avoidance. This is further clarified by Fig.~\ref{droplet_maneuver}, which shows the trajectory the vehicle will fly once an obstacle is detected (black dashed line), starting at the location indicated in the figure. Note that in this figure the coordinate system is defined with reference to the vehicle and that the camera system is mounted on the vehicle with a heading offset angle $\psi_{offset}$. This offset angle is defined as:

\begin{equation}
\psi_{offset} = \arcsin\left (\frac{R_{turn}}{CP_{dist}} \right )
\end{equation}

By introducing the offset, the avoidance maneuver can be performed in two steps, rather than three. First, the vehicle maintains its heading and flies to the turn point $TP$. Second, when $TP$ has been reached, the vehicle will perform a turn with constant rate and speed. This two-step procedure is visualized by the (black) dashed line in Fig.~\ref{droplet_maneuver}~(left). Note that it is assumed that the course of the vehicle is exactly the same as the heading angle. This assumption only holds if there is no crosswind and if the lateral drift is small. Also note that the vehicle performs only turns to the right. If $\psi_{offset}$ would be the same but negative, the vehicle would have to perform left turns.

If $\psi_{offset}$ would be zero or would have a different value, the heading angle of the vehicle would not be aligned with the direction of the turn point. An additional turn would then be required where the vehicle aligns with the circular path. This is shown in Fig.~\ref{droplet_maneuver}~(right). Thus $\psi_{offset}$ serves to reduce the complexity of the avoidance maneuver.

The time $\tau_{TP}$ needed to reach $TP$ is given by:


\begin{equation}
\tau_{TP} = \frac{TP}{V} = \frac{\sqrt{CP_{dist}^2-R_{turn}^2}}{V}
\label{eq:TP}
\end{equation}

Note that this timing value is only valid in case objects are detected as soon as they enter the Droplet area (perfect sensing). As explained in Sec.~\ref{detection_rules}, two adjustments were made to the detection algorithm to make it more robust in the real world. First, it is checked if there is sufficient texture present in the stereo images. If this is not the case, such observation is regarded unreliable and the system should respond similar as to the case when an obstacle is detected. Therefore the same timing value of Eq.~\ref{eq:TP} applies. The second adjustment takes into account distances to detected obstacles. For each point detected in the Droplet area it is computed how far it has penetrated this area. In other words, the distance between each detected point and the upper border of the Droplet area is computed. These individual distance estimates are used to obtain updated location estimates of $TP$ which are stored in the obstacle-detection counter array. By checking at each time step whether the value of the obstacle-detection counter that corresponds to the current vehicle location exceeds threshold $\tau_{d>ref}$, it is determined if a turn point has been reached. If that is the case, an avoidance maneuver is initiated.

\begin{figure}
\centering
  \includegraphics[width=0.38\textwidth]{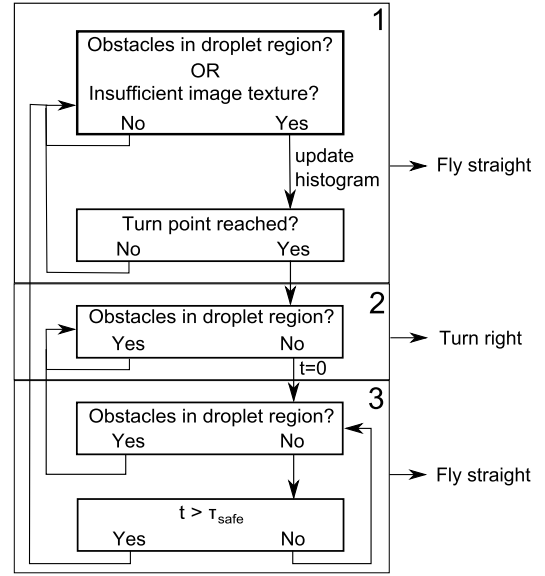}
  
\caption{The state machine for controlling the vehicle. The states define which outputs/actions are required, and include tests to check if a new state needs to be activated, or not. The value of $\tau_{TP}$ is precomputed, the value of $\tau_{safe}$ is a fixed tuning parameter.  }
\label{control_logic}       
\end{figure}

The diagram in Fig.~\ref{control_logic} shows the finite-state machine which is designed to ensure that the vehicle will always remain within the safety region. The first state will be active for as long as no avoidance action is required. Once the global time array indicates that a turn point has been reached (threshold $\tau_{d>ref}$ is exceeded), the second state activates. In this state the vehicle is instructed to perform the steady turn. The turn will continue until the vision system detects a new heading for which an obstacle-free Droplet region is found. If this is the case, the third state becomes active, and the vehicle will fly straight again. If the vision system suddenly detects obstacles while this state is still active, the second state immediately becomes active again and the vehicle is instructed to continue turning. If no obstacles are detected for a predefined amount of time (defined by the threshold $\tau_{safe}=$1~s) the system will return to the first state. 

The reason for adding the third state is twofold. First, due to inertia the heading angle response will have some overshoot when the vehicle stops turning. New obstacles might then be present on the right side of the Droplet region. Second, due to turning a higher level of motion blur is present in the camera images. Some obstacles are therefore only properly detected while flying straight. By adding the third state the chance of detecting obstacles is increased. This state can potentially be left out if the onboard control system would take care of overshoots and when more sensitive cameras would be used that suffer less from motion blur. 

As a final note it is emphasized that all parameters that can be set are fixed during flight. This holds for the vehicle dynamics, such as forward speed and turn rate, but with that all other settings that define the Droplet shape and timings. All aspects of the avoidance strategy are therefore fully precomputed which makes the algorithm extremely efficient.


\subsection{Theoretical guarantee of collision-free flight }

\begin{figure}[t]
\centering
    \includegraphics[trim={0.8cm 0 1cm 0cm}, clip,width=0.48\textwidth]{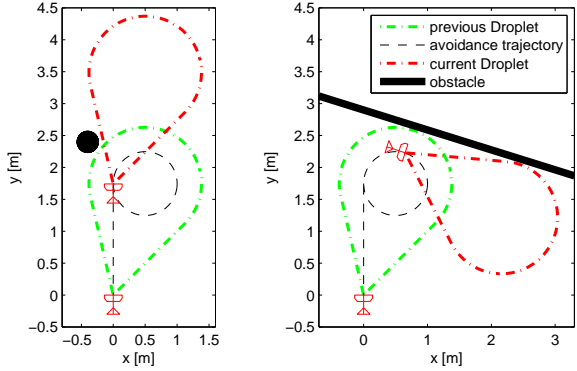}

\caption{Illustration of the working principle of the Droplet strategy.\protect\\ \textbf{Left}:~Example of the avoidance of a small round obstacle. In this specific case, the avoidance maneuver is triggered when the obstacle is detected within the lower droplet area (green). Once the vehicle reaches the turn point, the droplet area (red) is found to be free of obstacles, and the avoidance maneuver is aborted. \textbf{Right}: A more general example of the avoidance of a wall. In this case, the avoidance maneuver is triggered when the wall is detected within the lower droplet area (green). The vehicle will turn according to the predefined avoidance trajectory until the droplet area (red) is found to be free of obstacles again. }
\label{droplet_proof_2}       
\end{figure}

\begin{figure}[t]
\centering
    \includegraphics[width=0.48\textwidth]{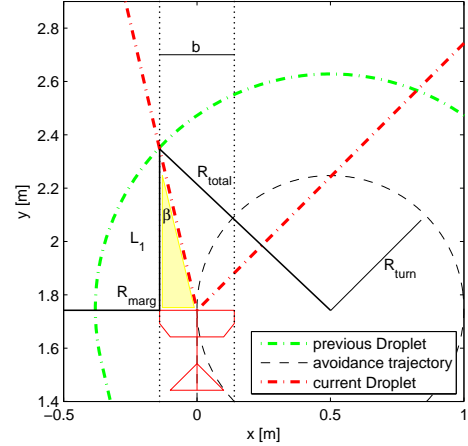}

\caption{ Visualisation showing that the vehicle always flies in free spaces that it has observed before. In the current situation the vehicle is enclosed by a previously observed Droplet region (green). It will be enclosed by the currently observed Droplet region (red) as soon as it leaves the previous (green) region along a tangent line of the turn circle.  } 
\label{droplet_proof}       
\end{figure}

The droplet strategy is guaranteed to avoid collisions, given perfect sensing and actuation. The main reason for this is that a robot employing the strategy will always move within free space that it has observed before, which is illustrated by two examples in Fig.~\ref{droplet_proof_2}. There are two conditions to this guarantee: (1) there is no obstacle to the front left of the robot at initialisation, and (2) the margin $R_{marg}$ is large enough. The reason for the first condition is illustrated by Fig.~\ref{droplet_proof}. The yellow (solid) triangle indicates a region right in front of the vehicle that is not covered by the field-of-view of the camera system. At initialisation there should be no obstacle in this region. 

The second condition ensures that for the remainder of the flight this same (yellow) region fits within an earlier observed free droplet region. This is true if $R_{marg}$ is of sufficient size. As defined in Fig.~\ref{droplet_proof}, the size of $L_1$ can be determined as:

\begin{equation}
L_1 = \frac{b}{2\tan\beta} =\frac{b}{2\tan(HFOV/2-\psi_{offset})}  
\end{equation}

Furthermore, $R_{total}$ can then be expressed as:

\begin{equation}
R_{total} = \sqrt{L_1^2+(b/2+R_{turn})^2}
\end{equation}

Using Eq.~\ref{radius}, a minimum value for $R_{marg}$ can then be obtained. This is the theoretical minimum value of $R_{marg}$, assuming no errors. If the actual value of $R_{marg}$ is larger than $R_{marg}^{min}$, the Droplet strategy guarantees collision-free flight. Since in the real world, sensing and actuation are not perfect, it is better to take a higher margin. Indeed, the Droplet size from Fig. 6 has $R_{marg}= $30~cm, which is larger than the theoretically required $R_{marg}^{min}$ of 21~cm.


\subsection{Comparison of computational complexity}


As stated in the introduction the proposed avoidance strategy is computationally efficient. To put this claim into perspective, the computational and memory complexity of the proposed method is compared to related methods. Different elements of such methods can be distinguished: sensors (with or without dedicated processor), post-processing steps (changing sensor data representation, e.g., making a map) and control algorithms (e.g., reactive, path planning, our Droplet method). In analogy to Section~\ref{related} most of the methods can be categorised as either reactive methods or path planning methods. Reactive methods are computationally extremely efficient, as they typically just compare some sensor values to pre-set thresholds, with a few if/else statements~\cite{zufferey2010optic} or a small neural network~\cite{ross2013learning}. However, as was mentioned, and as will be demonstrated in Section~\ref{simulations} by simulation experiments, reactive methods do not guarantee collision avoidance in certain situations. As explained in Section~\ref{avoidance_maneuver} the Droplet strategy requires a straightforward comparison between disparity maps generated by the stereo vision system and a precomputed reference disparity map. This is the only input for the simple state machine. The computational complexity of our approach is therefore in the same order as reactive methods.

As this study focuses on an obstacle avoidance strategy embedded into a real MAV, its computational complexity is compared to three path planning approaches that are intended for applicability to MAVs. The first approach we compare with uses stereo vision in combination with an RRT planner \cite{matthies2014stereo}. It requires three steps to compute a safe path: an expansion-operation on the disparity map to correct for the size of the MAV, the computation of dynamically feasible trajectories to randomly proposed waypoints using a closed-loop RRT algorithm, and a check whether candidate trajectories collide with the obstacles detected in the disparity map. The computational complexity is not specified in detail but it is mentioned that on a 1.86 GHz processor this method produces motion plans at 2Hz. Furthermore it should be noted that the disparity maps are generated by a separate stereo camera system with a dedicated processor. The second approach we compare with also uses a separate stereo vision system to produce disparity maps \cite{xu2015real}. This information is converted into a memory efficient octree-based search lattice. An $A^{\ast}$ graph search is used to find a collision-free optimal path to a goal state. For generating the motion plans based on the disparity map a 1.7 GHz processor is used. It uses under 30\% of CPU and about 400 MB of memory to deliver motion plans at approximately 2 Hz. The third approach we compare with uses a LIDAR to detect obstacles \cite{macallister2013path}. Using these measurements an occupancy grid is obtained which is used by the motion planner (variant of $A^{\ast}$) to find feasible trajectories based on motion primitives, taking into account the 3-D footprint of the vehicle. For their experiments on the real platform a 2 GHz processor is used. Using 60\% of CPU a motion plan is computed at 0.5 Hz on average. For comparison, it is noted once more that the obstacle avoidance system proposed in this study combines all steps from sensing to control decisions on a 4~g vision system that relies on a single 168~MHz processor with 192~kB memory, and still runs faster than 15 Hz. Hence, the memory required is more than 2000 times smaller than the approach of \cite{xu2015real} (400 MB / 192 kb = 2142) and $-$ if we take the processor speed at face value and assume 1 core used $-$ at least 75 times faster (1.7GHz / 168 MHz $\approx$ 10, 15 / 2 = 7.5). This is a very prudent estimate, since the numbers of the droplet strategy include the stereo vision processing time and memory, while those of [36] exclude the stereo processing information.

It is further noted that the three studies that use path planning also rely on additional sensors for pose estimation and localisation, either by running an onboard SLAM algorithm or by relying on external tracking. On the other hand, path planning methods perform a more sophisticated task than pure obstacle avoidance; they aim to arrive at a goal position. A downside of path planning methods in general is that accurate information of obstacle sizes and locations is required to compute safe paths around them. The method proposed in this study circumvents this need which makes it suitable for small scale systems with embedded sensing and processing.

\section{Simulation experiments}\label{simulations}

A simulation setup was created in \textit{SmartUAV}, a software environment which has been developed in-house, to compare the proposed avoidance strategy with other reactive strategies from the literature and to analyze effects of several parameters on its performance. The software simulates the motion of the vehicle (the DelFly in this case), the visual inputs to the stereo vision camera, the stereo vision algorithm (described in Section~\ref{vision}) and the Droplet control loop. The motion of the vehicle is simulated at 50 Hz, the vision and control loops are simulated at 10 Hz. 

Because the performance of the avoidance strategy is the point of interest in the simulations, two external factors of influence are ignored. First, no wind disturbances are taken into account because the platform currently does not have the ability to estimate this factor. Second, the simulations are performed in a highly textured environment such that the performance of the stereo vision algorithm is reliable and constant. These simplifications allow for a good comparison of different avoidance strategies. 

\begin{figure}[t]
\centering
  \includegraphics[width=0.485\textwidth]{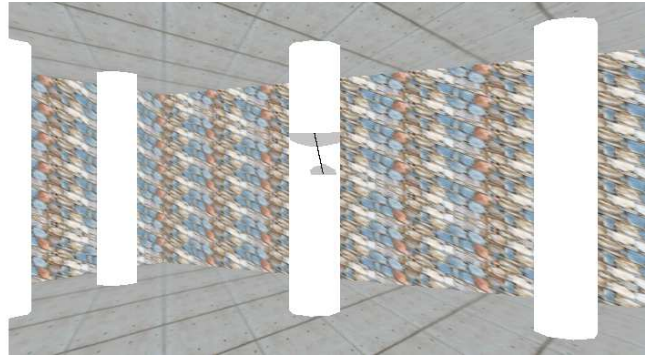}
\caption{Screenshot of the SmartUAV simulator that shows a DelFly model flying in a simulated room. In this example the walls are highly textured, the vertical poles are white and have no texture (for visibility of this image), the floor and ceiling are visualized as concrete stones.  }
\label{smartuav}       
\end{figure}

In the setup, the DelFly is flying in a square room of 6$\times$6$\times$3 meter with textured walls as shown in Fig.~\ref{smartuav}. Every single run, five obstacles, having the same texture as the walls (white in the figure for visibility), are randomly located to increase the difficulty of the avoidance task. Note that this is a very challenging environment, since it is a relatively small, closed space with additional obstacles. The obstacles are vertical poles with a diameter of 40~cm and a height of 3~m. It is assured that the DelFly starts in a position where it will detect an obstacle-free Droplet region. Each experiment run ends as soon as a crash occurs or stops after 600 seconds of uneventful flight. This time limit represents the maximum flight time with a single battery. The DelFly flies at a constant height of 1.5~m and with a constant velocity of 0.55~m/s. 



\subsection{Comparison with other Reactive methods}

The Droplet strategy is compared to two other strategies. The first one is based on the method proposed by \cite{zufferey200710} and \cite{moore2014autonomous} which aims to balance the average optical flow as measured by cameras on the left- and right-hand side of the vehicle. Since the vision system in our study obtains a single disparity map instead of optical flow from two different cameras, the disparity map is split into a left and right half, and the average disparity values of the two halves serve as the input that needs to be balanced. The method often resulted in crashes in one of the corners, since the walls act as a 'funnel' in these cases. A constant turn rate offset (18\% of maximum turn speed 120~$^{\circ}$/s) is therefore added which results in successful flight when the room is free of obstacles. The offset is chosen such that the resulting behavior is comparable to the results described in \cite{zufferey200710}. It was further tuned during simulations to obtain the best performance results for the eventual flights with obstacles. This method is referred to as the \emph{balancing} method.

The second strategy is based on a method proposed in de~Croon $et~al.$ \cite{DECROON2012C}. This method is based on time-to-contact estimates obtained from optical flow. Based on time-to-contact estimates from the left and right halves of the camera images, it is determined whether the vehicle should start turning, and in which direction. Once a turn has been initiated, this turn is continued for a fixed amount of time to avoid oscillations. In this study we define a disparity threshold (4~px) for both halves of the disparity map. If sufficient pixels ($\geq$10) in one of the halves of the disparity map exceed the threshold, the vehicle starts turning with maximum turn rate (120~$^{\circ}$/s) in the direction opposite of the detected obstacle. In our implementation a new turn direction can only be chosen once the vehicle flies straight again. This method is referred to as the \emph{left-right turning} method.

\begin{figure}[t]
\centering
  \includegraphics[width=0.485\textwidth]{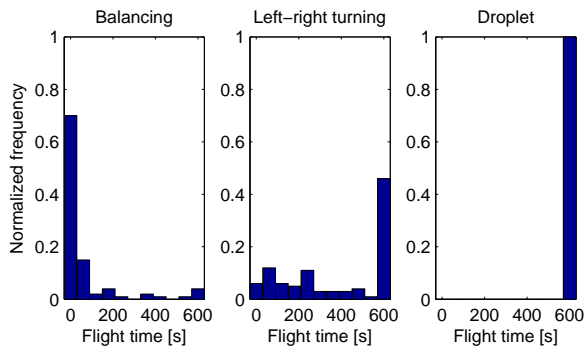}
\caption{Comparison of performance of three different reactive avoidance methods. The methods are indicated as Balancing method \cite{zufferey200710}, left-right turning method \cite{DECROON2012C} and Droplet method proposed here. The histograms visualize flight time until an obstacle is hit, with a maximum of 600 seconds. The results are normalized and are based on 100 runs for each method.}
\label{histograms}       
\end{figure}

\begin{figure}
\centering
  \includegraphics[trim={1cm 0 1cm 0}, clip, width=0.485\textwidth]{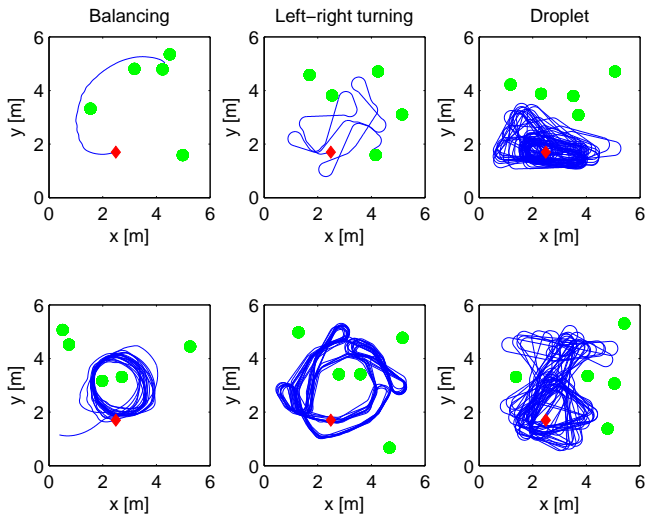}
\caption{Example flight trajectories showing the behaviors of the three avoidance methods in a room of 6$\times$6 meters containing five obstacles (indicated by the green circles). The red rhombus indicates the starting position. A failed flight (top-left) and a successful flight (bottom-left) of the Balancing method are shown. The two middle graphs show a failed flight (top) and a successful flight (bottom) of the left-right turning method. The two right graphs show two successful flights of the Droplet method proposed in this study. These trajectories show how the locations of the obstacles affect the areas covered by the vehicle.     }
\label{tracks}       
\end{figure}

For this comparison the Droplet strategy has the dimensions as shown in the top-left of Fig.~\ref{param_changes}. The value for $R_{turn}$ is 263~mm, which follows from the selected velocity (0.55 m/s) and turn rate (120~$^{\circ}$/s).

Fig.~\ref{histograms} shows the distribution of flight times of 100 runs for the different strategies in separate normalized histograms. The first histogram indicates that the balancing strategy fails in most cases. For the implementation in the original study \cite{zufferey200710}, the system was tuned for an empty room. Neither the vision system nor the control strategy was supposed to cope with other obstacles in the test room. The two images on the left in Fig.~\ref{tracks} show the tracks of a failed flight and a successful flight using this method. The track of the successful flight demonstrates that a safe route is found for certain obstacle setups. The track of the failed flight is more representative for the general behavior, however. It demonstrates that relatively small obstacles influence the flown trajectory but not sufficiently to steer the vehicle away from too narrow passages. 

The left-right strategy has a much higher success rate. The two middle plots in Fig.~\ref{tracks} show the tracks of a failed flight and a successful flight. The plot of the failed flight shows a typical failure case for this method. Due to the limited camera field-of-view, an avoidance turn is initiated in the direction of an unobserved obstacle. The track of the successful flight shows that this method allows the vehicle to reach a large part of the room. It can also be noted that the resulting flight trajectories are repetitive, which is typical.

Fig.~\ref{histograms} shows that the method proposed in this study results in a 100\% success rate, as expected. The two plots on the right in Fig.~\ref{tracks} illustrate that the region of the room covered during the flight strongly depends on the obstacle locations. It can also be observed that the flight tracks are less repetitive when compared to the other strategies.

\subsection{Parameter variation} \label{parameters}

As indicated by the shape parameters in Equations~\ref{eqn:width} and \ref{eqn:length}, the shape of the avoidance region is defined by the total radius $R_{total}$ and the horizontal field-of-view angle $HFOV$ of the camera system. The size of the total radius depends on the turn radius $R_{turn}$ which is determined by the selected values for forward flight speed $V$ and turn rate $\dot{\psi}$. In this section the influence of these parameters on the shape of the Droplet area and the flight trajectories is analyzed. As a baseline, the same values as in the simulations from the previous section are used: a forward velocity of 0.55~m/s and a turn rate of 120~$^{\circ}$/s. Again, the Droplet shape is defined as shown in Fig.~\ref{param_changes}(a). Three different cases of parameter variation are analyzed. 

\subsubsection{Higher forward speed and turn rate}

When the turn speed and turn rate are increased by the same factor (1.2), the shape of the avoidance region does not change. The time to reach the turn point $\tau_{TP}$ will decrease because the avoidance maneuver is flown at a higher speed. Since the turn rate is increased but the stereo vision update rate is not changed (fixed at 10~Hz for all simulations), the heading sampling angle between the stereo vision measurements during the turn increases. As a result, the vehicle will stay longer in turns as the chance of finding a safe heading is reduced. Fig.~\ref{param_changes}(d) shows an example where the vehicle is locked at a few turn locations for long periods. This result illustrates that for selecting the values of forward speed and turn rate, the update rate of the vision system is also important. 

\subsubsection{Lower forward speed}

By lowering the forward speed (0.36~m/s) while keeping the turn rate the same, the turn radius becomes smaller. This results in a decrease of the width, length and area of the avoidance region, which is visualized in Fig.~\ref{param_changes}(e). As a result the vehicle is able to access smaller spaces and the coverage in the simulated room should increase. Fig.~\ref{param_changes}(f) shows a flight trajectory which is an example of a flight where the vehicle is indeed able to cover multiple areas in the simulated room.

\subsubsection{Larger field-of-view}
\begin{figure}
\centering
  \includegraphics[trim={0cm 1cm 0cm 1cm}, clip,width=0.485\textwidth]{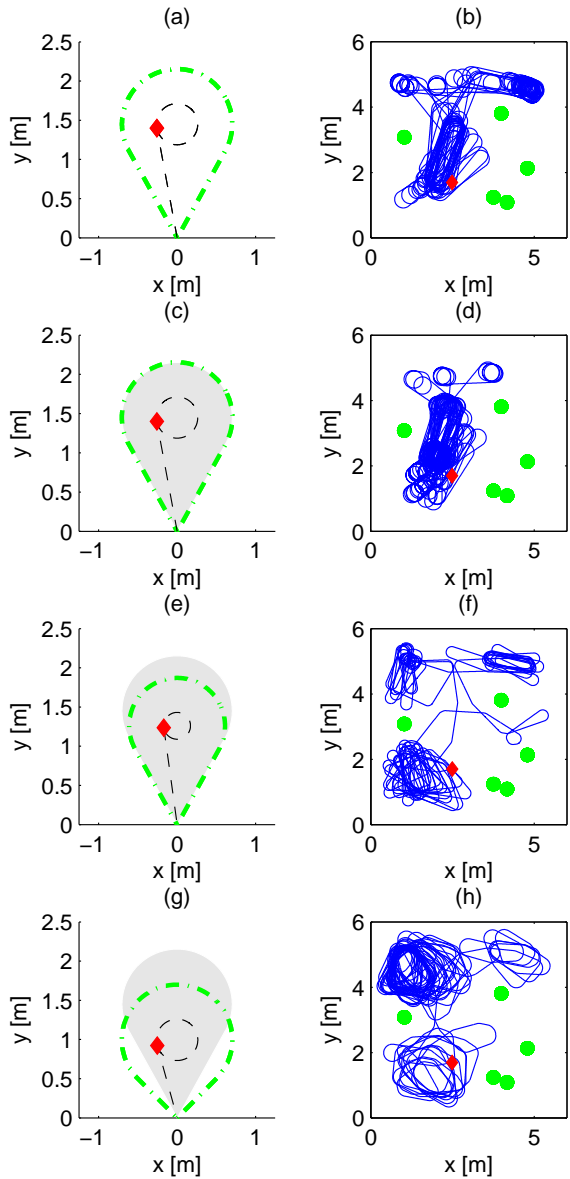}
\caption{Shape of the Droplet region for different parameter settings (left column) and example trajectories for these Droplet shapes (right column). They gray regions indicate the shape as defined in Plot \textbf{a}, which is the shape of the baseline configuration. Plots \textbf{c-d} correspond to the case of higher forward speed and turn rate, Plots \textbf{e-f} are for lower forward speed and constant turn rate, and Plots \textbf{g-h} are for a larger field-of-view. The room layout is identical in the four cases to provide a clear visual comparison.}
\label{param_changes}       
\end{figure}

If only the horizontal field-of-view is increased (from 60$^{\circ}$ to 90$^{\circ}$), the width of the avoidance region does not change, but the length and area are decreased as shown in Fig.~\ref{param_changes}(g). This should also have the effect that smaller spaces can be reached and that the coverage increases. Fig.~\ref{param_changes}(h) shows a flight trajectory for this configuration that illustrates that the vehicle is able to reach different parts of the simulated room. 

Table~\ref{table} shows statistics for the varying parameter configurations based on 200 runs per parameter setting. These results show that flying at higher speeds and turn rates has the effect that the vehicle will be turning more often. Note that the number of turns is almost the same, but that the mean time per turn is increased. It can also be seen that this has a negative effect on the total coverage. Using the bootstrap method \cite{cohen1996empirical} it is determined that this decrease is significant ($p$\textless0.01).

Flying at a lower speed (but with the same turn rate) and thereby reducing the size of the Droplet region leads to an increase in the area covered during a flight. This increase is also found to be significant with $p$\textless0.01. The amount of time spent on turning is decreased. On the other hand, the number of turns increases. As a result, the time per turn is decreased on average. These facts indicate that by lowering the flight speed, the vehicle makes more but shorter turns and is able to reach more places.

Increasing the horizontal field-of-view angle does not lead to a significant increase in area covered ($p=$~0.32). Apparently the width of the avoidance region is a more crucial factor than the size of the field-of-view. The mean time spent on turning reduces ($p=$~0.05) and the total number of turns increases. This can be explained by the reduced distance to the turn point ($CP_{dist}$), which leads to a reduced total time needed to perform the Droplet maneuver. More separate avoidance maneuvers are performed with smaller turn angles, which reduces the mean time turning and the mean time per turn.

\begin{table}[t]
\renewcommand{\arraystretch}{1.3}
\caption{Comparison of behavior of the Droplet method for different parameter settings}
\centering
\begin{tabular}{l*{4}{c}} 
\hline\hline
  & Base & 1) & 2) & 3) \\ 
\hline
Mean Coverage$^{1}$ (\%) & 52.3 & 44.9 & 58.0 & 53.8\\
Mean Time Turning (\%) & 38.2 & 49.4 & 24.4 & 35.1 \\
Mean Number of Turns & 118 & 117 & 138 & 270 \\
Mean Time per Turn (\%) & 0.32 & 0.42 & 0.18 & 0.13 \\
\hline\hline
\end{tabular}
\begin{tabular}{l}
\textbf{Base}: Reference parameter setting for the Droplet region as \\
defined in Fig. \ref{droplet_maneuver}. \textbf{1)} Higher forward speed and turn rate.\\
\textbf{2)} Lower forward speed. \textbf{3)} Larger camera field-of-view. \\
$^{1}$ Coverage is computed by dividing the test room in 36 patches \\
of 1$\times$1 meter and measuring how many patches are visited \\
by the vehicle at least once during a flight. Mean coverage is \\
computed as the average coverage over all flights. 
\end{tabular}
\label{table}
\end{table}


\subsection{Effect of extended obstacle detection rules}
As described in Sec.~\ref{detection_rules} different obstacle detection rules have been implemented. The standard implementation assumes that all obstacles are richly textured, such that obstacle detection using the stereo vision system is fully guaranteed. Two additional rules were described to deal with situations where objects might not be detected robustly. The first rule keeps track of the amount of observed texture to deal with poorly textured surfaces. The second rule implements a time history of obstacle detections, such that the chance of detecting a poorly textured object increases. 

The effectiveness of the additional obstacle detection rules is demonstrated by experiments in which the texture of the walls and the poles in the simulated room are varied. The performance difference between the standard rule and the additional rules is compared in terms of success rate and room coverage. The results are listed in Table.~\ref{texture_variation}. The left column indicates what is different compared to the fully textured room. Coverage is again expressed as percentage of visited area, averaged over all flights. The success rate indicates which percentage of 100 runs resulted in 600 seconds of collision-free flight.

An important observation is that the applied detection rules are very effective to prevent collisions with poorly textured surfaces, at the cost however of lowering the area covered by the vehicle. The success rates increase considerably for all conditions with poor texture. An interesting observation is that the covered area increases when the poles are not textured. Apparently, the vehicle visits locations that are never reached in case all obstacles are perfectly detected. In other words, the system takes more risk, as is reflected by the success rate which is not 100\% in this case.

\begin{table}[t]
\renewcommand{\arraystretch}{1.3}
\caption{Comparison of performance of obstacle detection rules}
\centering
\begin{tabular}{l|c c|c c} 
\hline\hline
      Rules: & \multicolumn{2}{|c|}{Standard}  & \multicolumn{2}{|c|}{Additional}    \\
      (\%) & Cov. & Suc. & Cov. & Suc.    \\ 
\hline
Fully textured & 46.5 & 100 & 37.8 & 100\\
One white wall & 45.3 & 65.0 & 37.4 & 100 \\
Two white walls & 26.7 & 25.0 & 18.8 & 100\\
Four white walls & 0 & 0 & 5.1 & 100 \\
White poles & 56.5 & 60.0 & 51.2 & 95.0 \\
\hline\hline
\end{tabular}
\begin{tabular}{l}

\end{tabular}
\label{texture_variation}
\end{table}

\section{Real-world Flight Experiments} \label{experiments}

Several flight tests with the real DelFly Explorer have been performed to evaluate the Droplet avoidance strategy. First, a set of eight flight tests was conducted that mimics the scenario from the simulation experiments. Furthermore, several flight tests were performed in different unadapted real world rooms to show the robustness of the method in all kinds of situations.  

\subsection{Experiments in simulator-like environment}

\begin{figure}
\centering
  \includegraphics[width=0.4\textwidth]{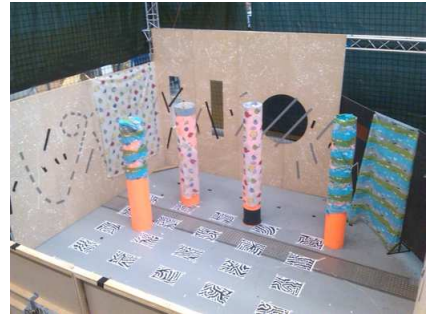}
\caption{Test room of 6$\times$6 meters where flight tests have been performed. The room contains four poles of 40~cm diameter that form obstacles. The locations of these obstacles were different for each test flight. }
\label{test_room}       
\end{figure}

A set of eight flight tests has been performed in a scenario that is comparable to the setup of the computer simulations. Fig.~\ref{test_room} shows the test location where walls are placed to form a closed square room. The tests mimic the simulation tests from Section~\ref{simulations}: the room measures 6$\times$6 meters, it is well-textured and contains four highly-textured obstacles at varying locations. The location of the vehicle is tracked using an OptiTrack\footnote{\url{http://www.optitrack.com/}} motion capture system. The vehicle is fully autonomous during the test flights. The altitude is controlled using feedback from the pressure sensor, heading is controlled based on the data from the stereo vision camera. It is important to note that in these experiments the obstacle detection adjustment rules are not implemented. These adjustments deal with situations where low-texture forms a problem for detection. By assuring good texture everywhere these situations do not occur in this experiment. Roll rate and pitch rate feedback from the gyroscopes is used to stabilize the heading and pitch angle. The flight speed ($\approx$0.6~m/s) is not regulated but set at the start of the flight by the trim position of the elevator. This elevator/speed setting allows for flight times of up to 9 minutes. A trim value for the motor rpm is set to minimize vertical speed. This trim value is set higher for the case a turn is made. Due to aileron deflections while turning, extra drag behind the wing is generated, which lowers the effective thrust. The higher setting for motor rpm mostly compensates for this loss during turns. 

The flight speed and avoidance turn rate have been tuned to obtain a good avoidance performance during the test flights. The corresponding Droplet shape is as shown in Fig.~\ref{droplet_maneuver} which has a length of 2.9 m and a width of 1.9 m. It is larger than the Droplet region indicated as baseline in the simulations (length of 2.1 m), because a slightly higher flight speed is set to have longer flight times. To keep the turn rate the same, a larger turn radius is then needed. In fact, also a lower turn rate is selected, resulting in a turn radius ($R_{turn}$) of 0.5 m. The error margin ($R_{marg}$) is the same, 0.3 m. The lower turn rate is necessary to prevent the vehicle from staying in lengthy turns. Such a situation can occur if there is only a small margin available for fitting the droplet region in between the poles. Due to the combination of a high turn rate with a limited frame rate of the vision system, the chance of observing a free droplet region is fairly limited in some situations. This effect is reinforced by a rule in the second state of the state machine that checks if a free droplet region is observed in two consecutive frames. This rule is meant to improve robustness of the second state as the turning motion of the vehicle improves the blurring effect in this state, which obviously leads to more false-negatives of the stereo matching algorithm. An additional cause for the lengthy turns can be an overshoot in final heading angle after ending a turn and switching to the third state. As a result new obstacles might come into view, making the state machine to switch back to the second state. 

These considerations make clear that the choice for the parameters in Eq.~\ref{turn_radius}, which are forward speed, heading turn rate and turn radius, are important to obtain robust performance. This is especially true for the conditions in the test room which is relatively small and cluttered. A high forward velocity is desired for long flight times and a small turn rate is desired to obtain a good performance from the stereo camera during turns. This leads to large values for the turn radius and thus to large droplet sizes. This is undesirable as the accuracy of the stereo vision system degrades with distance (worse obstacle detection) and because this also results in lengthy turns, as discussed previously. Long turning times have led to several collisions due to random drifting of the vehicle over time. The selection of the turn parameters from Eq.~\ref{turn_radius} turned out to be the most important step to obtain robust performance. Apart from this it was experienced that trimming and tuning the vehicle in between test flights is also crucial. The trim setting of the aileron is important to minimize the drift of the heading angle, which is necessary to fly straight when executing the first part of the droplet maneuver. Several crashes occurred in cases where the vehicle would drift to the right during the first phase, resulting in crashes while executing the turn maneuver. The aileron trim setting also affects the induced drag caused by the aileron, which affects the response of the vehicle during a turn. Too wide turns were another reason for several crashes. Therefore, tuning of the aileron turn command is required to obtain the desired turn rate. Experience shows that by careful construction the aileron system can be made more robust, but will degrade after long flight times. Note that tuning the Droplet parameters only needs to be done once for a certain vehicle configuration. Trimming the aileron offset and the aileron turn command needs to be done also in between flights. Future work will focus on making the system more robust in this sense. 

Robust performance is obtained given that the previously discussed conditions are satisfied. Furthermore, external factors such as the air flow and the lighting conditions play an important role in the robustness of the algorithm. Due to the relatively low wing loading of this specific vehicle it is important to have stable wind conditions, as drafts make the vehicle drift away from the intended (avoidance) trajectory. Several crashes occurred in cases where a sudden draft was experienced. Additional feedback on the position/velocity of the vehicle would be required to make the system more robust to this influence. This could be achieved by tracking surrounding objects using an additional algorithm in the stereo camera system, or by using an additional sensor. The quality of the disparity maps produced by the stereo camera system is obviously influenced by the lighting conditions since the camera sensors are passive. It was observed that daylight conditions resulted in a far more stable performance. By turning off artificial lighting or by testing without external sunlight, crashes would often be caused by a poor obstacle detection performance. Furthermore, it was mentioned previously that the turn rate of the vehicle has an important effect on the blurring of the camera images. The robustness of the method cannot be guaranteed when the camera system cannot produce reliable estimates. The limitations of the vision system might improve in the future as camera technology progresses.



\begin{figure}[t]
\centering
  \includegraphics[trim={1.5cm 0 1cm 0}, clip,width=0.47\textwidth]{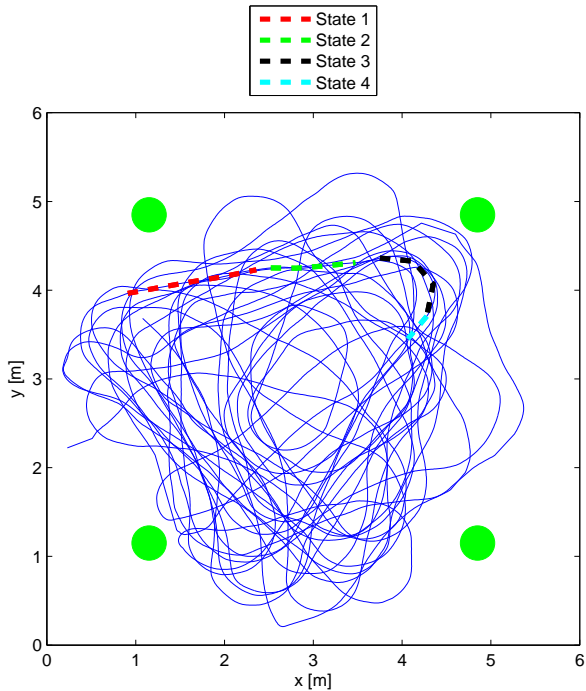}
\caption{Flight trajectory from Test $5$. During this flight of 410 seconds the vehicle covered a distance of 271 meters. The locations of the obstacles (40~cm diameter) are indicated using the (green) circles. The dashed line shows one complete cycle of the Droplet strategy; the sub-track starts in State 1 (directly after a turn) and stops after completing the next turn. }
\label{flight_track}       
\end{figure}

The flight trajectory of Test 5 is shown in Fig.~\ref{flight_track}. Part of the trajectory is indicated by the dashed line. This part of the flight starts immediately after finishing a turn, which means that State 1 is active. Further on, States 2, 3 and 4 become active, respectively. It can be observed that during this maneuver the DelFly keeps a safe distance to the obstacle it approaches. Other parts of the flight track show that during other approaches of this pole the vehicle gets much closer. This effect can be explained by the sensitivity of the DelFly to small gusts (non-constant flight speeds) and the degraded performance of the stereo vision system during flight. 

Results from the test flights are shown in Table~\ref{table2}. These results correspond to successful flights with different obstacle locations. Only successful flights are recorded that were achieved after obtaining good trim settings. The $Number~of~Turns$ is shown twice; the actual amounts of turns that were performed during the flights are shown, as well as the time-corrected numbers of turns, such that the test flights (with varying flight times) can be compared to the results from the simulated flights of 600~seconds as shown in Table~\ref{table}. Note that during flight Test $2$ and $3$, the DelFly performed over a hundred turns autonomously while flying in a 6$\times$6~m space with several obstacles. The covered area in the real test flights is higher which can be explained by the number of obstacles being four instead of five. Note that the $Time~per~Turn$ found for the flight tests with the longest durations (1-6, 8) matches the result found for the baseline case in Table~\ref{table}. The large variation in total flight time can be partly explained by the trim setting of the aileron that was tuned at the start of each flight. A small difference in trim setting can have a significant effect on the drag it creates.

\begin{figure}[t]
\centering
  \includegraphics[trim={1.5cm 0 1cm 0}, clip,width=0.47\textwidth]{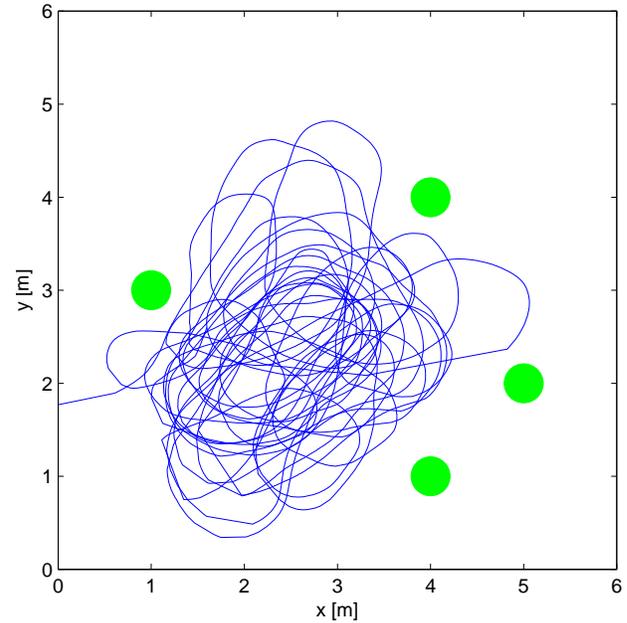}
\caption{Flight trajectory from Test $8$. During this flight of 295 seconds the vehicle covered a distance of 178 meters. The locations of the obstacles correspond closely to the obstacle locations in Fig.~\ref{param_changes}.  }
\label{flight_track_4}       
\end{figure}

In Fig.~\ref{flight_track_4} the flight trajectory of Test $8$ is shown. The obstacle locations in this test are very similar to Fig.~\ref{param_changes} to compare the flight track from the real flight with the computer simulation. It can be observed that the lower part of the room is better covered during the real flight than in the simulation which is caused by the more random nature of the DelFly behavior in real life. The upper part of the room is not visited by the DelFly during the real flight test. This can be explained by the larger size of the Droplet region in the test flights.


\begin{table}[t]
\renewcommand{\arraystretch}{1.4}
\caption{Flight trajectory results from real test flights}
\begin{tabular}{p{2.1cm} p{0.3cm} p{0.3cm} p{0.3cm} p{0.3cm} p{0.3cm} p{0.3cm} p{0.3cm} c} 
\hline\hline
  & 1 & 2 & 3 & 4 & 5 & 6 & 7 & 8 \\ 
\hline
Coverage (\%) & 69 & 64 & 75 & 64 & 72 & 64 & 58 & 61 \\
Time Turning (\%) & 41 & 37 & 41 & 53 & 40 & 46 & 45 & 47 \\
Nr. of Turns & 85 & 119 & 122 & 90 & 94 & 72 & 38 & 76\\
\emph{Nr. of Turns$^{\star}$} & \emph{127} & \emph{130} & \emph{136} & \emph{175} & \emph{138} & \emph{128} & \emph{89} & \emph{154}\\
Time / Turn (\%) & 0.32 & 0.28 & 0.30 & 0.30 & 0.29 & 0.36 & 0.50 & 0.31\\
Flight time (s) & 400 & 550 & 539 & 308 & 410 & 336 & 256 & 295\\
Trav. distance (m) & 250 & 337 & 344 & 192 & 271 & 203 & 154 & 178  \\
Avg. speed (m/s) & 0.62 & 0.61 & 0.64 & 0.62 & 0.66 & 0.60 & 0.60 & 0.60 \\
\hline\hline
\end{tabular}
\begin{tabular}{l}
$^{\star}$ The number of turns indicated is scaled to a total flight time of \\
600 seconds\\
\end{tabular}
\label{table2}
\end{table}

\subsection{Effect of extended obstacle detection rules}

\begin{figure}
\centering
  \includegraphics[width=0.4\textwidth]{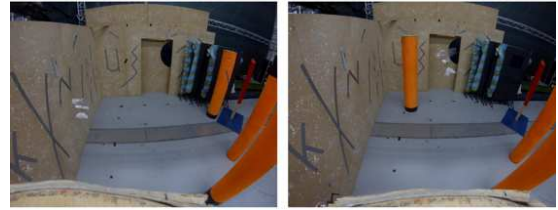}
\caption{Images of two setups to test the robustness of the system to poorly textured obstacles. The orange poles do not have additional texture. \protect\\ \textbf{Left:} room with two walls, enclosed by the orange poles. \textbf{Right:} same room as in the left case, one pole is placed inside the enclosed area.  }
\label{test_cyberzoo_poles}       
\end{figure}

The results from the previous tests demonstrate the performance of the Droplet method in case the poles and the walls in the room are well-textured. To test the influence of the extended obstacle detection rules, which take into account situations where objects are poorly textured, similar experiments are performed where no additional texture is used for increasing the visibility of the orange poles. As illustrated by Fig.~\ref{test_cyberzoo_poles}, a small setup is created where a part of the original test room is used which is demarcated by the poles. In the first experiment (left image) an open test space is created, in the second experiment (right image) one pole is placed inside this test space to increase the complexity of the avoidance task.

\begin{figure}
\centering
  \includegraphics[width=0.4\textwidth]{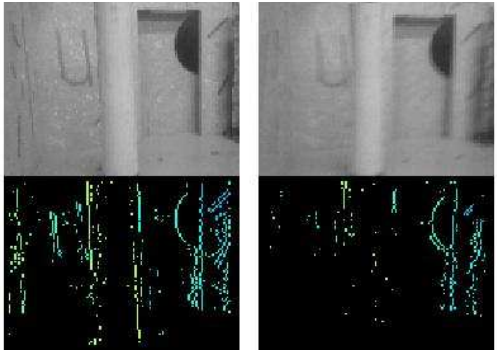}
\caption{Camera images and corresponding disparity images of the same scene under different camera conditions. The examples show the difficulty of detecting the orange poles, which look like bright vertical beams (top images). \protect\\ \textbf{Left:} camera is static. \textbf{Right:} camera is attached to the flying vehicle, thereby introducing motion blur.  }
\label{pole_detection}       
\end{figure}

To illustrate the complexity of the task and the challenge for the avoidance algorithm in this particular setup, Fig~\ref{pole_detection} shows examples of stereo vision output. The figure illustrates that the distinctive orange color of the poles does not result in a distinctive intensity difference in comparison with the background. The same scene is shown for two conditions, one while the camera is static (left image), and one while the camera is carried onboard the DelFly Explorer in flight (right image). These examples make clear that due to motion blur in the images, the resulting disparity maps can be affected significantly. The proposed detection rules take into account that the number of points in the disparity maps is considerably reduced and varies more due to motion blur. In general a useful number of points is still detected, making the detection method very effective. 

Results from flight test in both setups from Fig.~\ref{test_cyberzoo_poles} demonstrate that the system is able to robustly detect and avoid the orange poles when using the extended detection rules. Flights of $8$ minutes and $4.5$ minutes were recorded respectively. Videos of these flights are available in the video playlist\footnote{\url{https://www.youtube.com/playlist?list=PL_KSX9GOn2P987jTwx4szhPUPzpW5WJ3k}} that belongs to this paper. Tests without the extended detection rules result in many crashes and no successful flights. Especially in cases where an orange pole is present to the right of the vehicle, the chance of a detection failure turns out be too high to allow successful collision-free flights of several minutes.


\subsection{Experiments in unadapted real word environments}

\begin{figure}
\centering
  \includegraphics[width=0.4\textwidth]{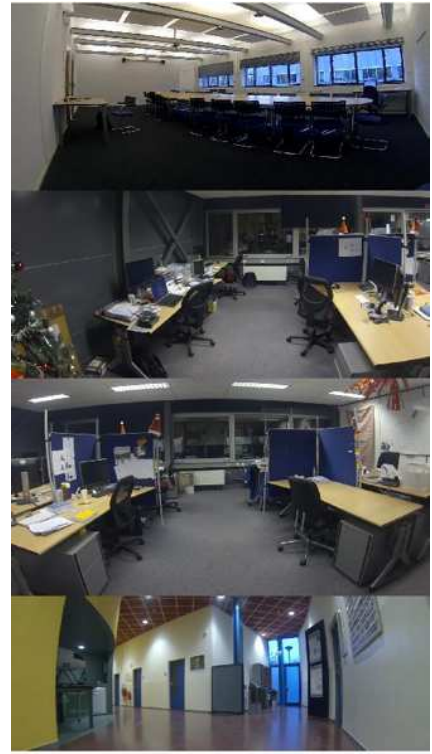}
\caption{Images of three rooms where real flight tests were conducted.\protect\\ \textbf{Top:} meeting room. \textbf{Two in Middle:} office spaces.  \textbf{Bottom:} Hallway.  }
\label{test_rooms}       
\end{figure}

Real flight tests were also performed with the DelFly Explorer in several unadapted rooms. Pictures of these rooms are shown in Figure \ref{test_rooms}. The first room is a meeting room with a long table with chairs in the middle. It contains three large white walls with little texture and one side with transparent windows. The second and third room are office spaces with screens surrounding the desks that form vertical obstacles in the middle of the rooms. The fourth space is a hallway with different features: curved walls, corners, corridor structures and several texture-poor walls. 


In all tests, the DelFly was able to fly for as long as the battery permitted. Videos of these flights are available in the video playlist. Different flight behaviors were found in every room. In the meeting room the vehicle had the tendency to follow the contours of the room. In cases where the vehicle would get further away from the wall, the table or chairs were often interpreted as obstacles, making the vehicle to turn sooner and crossing the room. In this room the distance between the ceiling and the table, being less than two meters) is relatively small. Due to the variations in altitude caused by barometric pressure differences and small gusts due to a climate control system incorporated in the beams of the ceiling, this situation occurs very often. For this reason the operator manually corrected altitude errors to allow testing the robustness of the Droplet strategy in the horizontal plane. A flight time of 7 minutes was recorded in this room. In the two office spaces the amount of space between the desks is relatively narrow. This causes the vehicle to perform many turns, which results in a considerable reduction of flight time. Nevertheless, flight times of over 3 minutes were recorded in these rooms. It can be observed that the videos always stop when the battery gets empty. In this specific environment this sometimes causes the vehicle to descent and hit an object with its tail. The hallway allows the vehicle to fly more straight parts, similarly to the meeting room, allowing also longer flight times. In this space the vehicle virtually bounces between the walls. Because the vehicle always makes right turns, it stays within the space that is visible in the image. In some cases the vehicle flies into the corridor on the left, but it always turns around at some point when approaching the wall. Flight times of over 5 minutes have been recorded in this environment. It can be observed that the vehicle may lightly touch a wall during an avoidance maneuver. This generally forms no problem for flapping wing MAVs. These situations can be prevented by applying trajectory following (instead of only timing the turn point) or by increasing the safety margin parameter of the Droplet area, but this makes the area larger and hence leads to reduced coverage. 



\section{Conclusions} \label{conclusions}

In this study a strategy is proposed for obstacle avoidance in small and cluttered environments, that takes into account both sensor limitations and nonholonomic constraints of flapping wing MAVs. The method relies on measured distances to obstacles from stereo vision. Its computational complexity, both in terms of time and space, make it specifically suitable for use on small, embedded systems. Simulation experiments show that the method ensures collision avoidance in a small and cluttered room. Real flight experiments demonstrate that the method allows a 20~g flapping wing vehicle to autonomously perform sustained flight of up to 9 minutes while avoiding obstacles and walls in different environments.

\appendix[Extending the avoidance maneuver to 3D]

\subsection{Vertical maneuver definition and implementation}

\begin{figure}[ht]

  \includegraphics[trim={2cm 3cm 4cm 1cm},clip,width=0.48\textwidth]{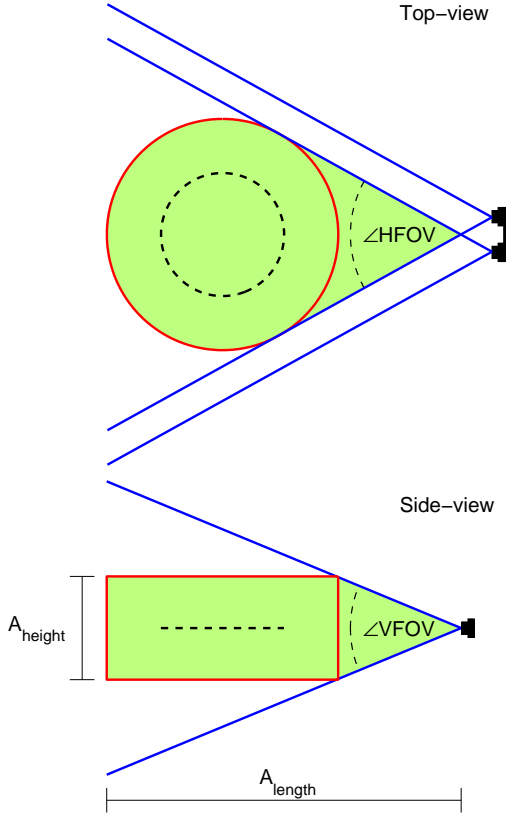}
\caption{Top view and side view of the droplet region. The green region needs to be checked for obstacles. The turn region, indicated by the red lines is defined as a disk with height $A_{height}$. for clarity the horizontal ($HVOF$) and vertical ($VFOV$) field-of-view angles of the cameras are indicated.  }
\label{droplet_hor_ver}       
\end{figure}

The Droplet maneuver can be extended into 3D by taking into account the ability of the vehicle to climb and descent. The vertical speed of the vehicle can be controlled independently from the horizontal states and is determined by the motor speed, which regulates the flapping frequency. Fig.~\ref{droplet_hor_ver} shows the droplet area (green) both from the top as well as from the side. From the side-view it can be seen that the camera vertical field-of-view ($VFOV$) plays a role in how this region is defined vertically. The height of this area is restricted to $A_{height}$  which is defined as:

\begin{equation}
\begin{aligned} 
A_{height} = h + 2H_{marg} \\ 
&\quad \\
\end{aligned}
\label{eqn:height}
\end{equation}

Here, $h$ is the vertical size of the vehicle and $H_{marg}$ is an error margin to account for obstacle detection inaccuracies and for altitude variations. The latter is determined by the performance of the height control loop. The red box indicates the region that must be kept clear of obstacles to allow a safe horizontal avoidance maneuver. It corresponds to the red circle in the top-view; its width is equal to the radius of the red circle. 

\begin{figure}[ht]

  \includegraphics[trim={2cm 3cm 2cm 1cm},clip,width=0.48\textwidth]{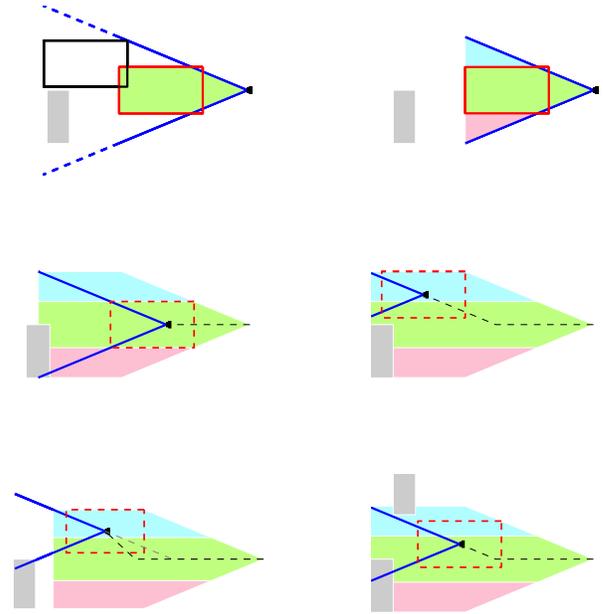}
\caption{Possible implementations of height control into the Droplet avoidance strategy. The gray boxes indicate obstacles that need to be avoided. \textbf{Top-left}:length of the droplet is extended such that the future turn region (black box) fits when shifted vertically. \textbf{Top-right}: length of the droplet is not changed, but observations from the areas above and below the droplet area are stitched together. After checking for obstacles over a certain amount of distance (\textbf{middle-left}) a climb maneuver can be initiated. If an obstacle is then detected within the Droplet area (green), the climb is executed (\textbf{middle-right}). If the obstacle is further away (\textbf{bottom-left}), the climb will be initiated later, as soon as the obstacle is detected. The climb can be steeper in this case. If an obstacle is detected in the climb region (blue), the climb can only be executed for as long as the corresponding turn region allows.    }
\label{droplet_ver_implement}       
\end{figure}

The regions above and below the red turn region (side-view) are observed by the cameras as well. They are ignored for making horizontal avoidance decisions, but for initiating vertical maneuvers these regions need to be checked for obstacles. By doing so, it can be guaranteed that after a vertical maneuver the corresponding turn region will be shifted into a region without obstacles. A possible method to implement vertical maneuvers is therefore to extend the length of the droplet area, as shown in the top-left diagram in Fig.~\ref{droplet_ver_implement}. The diagram shows how the turn region can be moved upwards (black rectangle) to allow a climb. However, this method requires the length of the droplet area to be increased substantially (dashed lines), while the inaccuracy of the stereo vision system degrades significantly at larger distances.   

A more compact method is proposed as visualized in the other diagrams of Fig.~\ref{droplet_ver_implement}. The top-right diagram shows that the length of the droplet region does not need to be increased. The climb area (blue) and descent area (red) are indicated which fill up the remainder of the $VFOV$. By stitching observations in these regions together over time, these areas are stretched out. The middle-left diagram shows how the observed area has grown while the camera has moved to the indicated location. This is a specific location; at this location the corresponding turn region (indicated by the dashed red box) is the same region as observed earlier in the starting situation as shown in the top-right diagram. The middle-right diagram illustrates that a climb maneuver can be followed from this point on. By following a climb path parallel to the border of the $VFOV$ it is guaranteed that the corresponding turn region (dashed red box) will always be enclosed within the region observed previously.

\begin{figure}[ht]

  \includegraphics[trim={2cm 3cm 0cm 2cm},clip,width=0.48\textwidth]{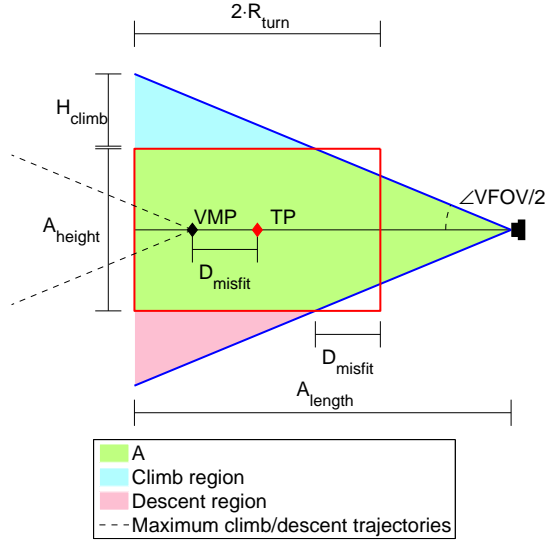}
\caption{Side-view of the Droplet avoidance area. The filled inner (green) area defines the region within the field-of-view of the cameras (indicated by the black symbol at the right) that needs to stay clear of obstacles. The length of this area, $A_{length}$, corresponds to the definition of Fig.~\ref{droplet_params}. The filled outer (blue) areas define regions that need to be clear of obstacles to allow a climb or descent maneuver. Such a maneuver is only safe to execute after the vehicle has travelled at least up to the vertical maneuver point $VMP$. The maximum rate for a climb/descent is limited by the field-of-view of the camera and the maximum climb/descent distance is defined by the field-of-view and $A_{length}$.   }
\label{droplet_vertical}       
\end{figure}

The method is further defined and generalized as presented in Fig.~\ref{droplet_vertical}. First note that in this example configuration, the red turn region is not fully enclosed by the $VFOV$. The size of the non-enclosed area is indicated as $D_{misfit}$, which is defined as:

\begin{equation}
D_{misfit} = A_{length}-\frac{A_{height}}{2\cdot \tan(VFOV/2)} \\ 
\label{eqn:distance_outside}
\end{equation}

Depending on the sizes of $A_{length}$ and $A_{height}$, the value of $D_{misfit}$ can also be zero or negative, meaning that the whole area is enclosed in the $VFOV$. $D_{misfit}$ is important as it determines at which point a vertical maneuver may be initiated. For clarity, we define the vertical maneuver point $VMP$. It is different from the turn point $TP$, which is used for the horizontal maneuvers. The relation between these two points is defined as:

\begin{equation}
VMP = TP + D_{misfit} \\
\label{eqn:vmp}
\end{equation}

In the configuration as presented in Fig.~\ref{droplet_vertical}, $VMP$ is further away than $TP$. The additional shift $D_{misfit}$ ensures that the turn region (red box) corresponding to $TP$ is shifted into the field-of-view before the vehicle starts climbing. 


When $VMP$ is reached, a vertical maneuver will be executed as soon as an obstacle is detected within the Droplet area. If an obstacle is detected as soon as $VMP$ is reached, the maximum climb/descent rate is limited by the slope of the $VFOV$ borders. This is shown in the middle-right diagram in Fig.~\ref{droplet_ver_implement}. If an obstacle is detected after reaching $VMP$, the vertical maneuver is also initiated later, and a higher climb/descent rate is allowed. This is shown in the bottom-left diagram of the same figure. The maximum climb/descent height difference $H_{climb}$, indicated in Fig.~\ref{droplet_vertical}, follows from the sizes of the $VFOV$ and the turn region. It is defined as:

\begin{equation}
\begin{aligned} 
H_{climb} = A_{length}\cdot \tan(VFOV/2)-A_{height}/2 \\ 
&\quad \\
\end{aligned}
\label{eqn:height_climb}
\end{equation}

Just like the horizontal avoidance maneuvers, the vertical maneuvers need to be planned ahead, based on the observations from the camera system. To incorporate height control, disparity observations need to be compared with two reference disparity maps. The first reference disparity map only represents the Droplet area (green), and is also used for the horizontal control decisions. The second reference map represents the whole $VFOV$ up to a distance $A_{length}$, thus also the spaces above and below this area (blue and red). This combination allows to identify whether obstacles are detected within the Droplet region or in the climb/descent regions. The finite-state machine from Fig.~\ref{control_logic} can be maintained to incorporate height control. For as long as the system is in the first state, it needs to be checked if the climb and/or descent regions are free of obstacles. No obstacles should be detected in these areas while travelling for a distance $D_{misfit}$ first. As long as this is not the case, the location of $VMP$ cannot be fixed, meaning that a vertical maneuver is not safe. If the climb or descent region is found to be free of obstacles for a distance $D_{misfit}$, the location of $VMP$ is fixed according to the definition in Fig.~\ref{droplet_vertical}. Thereafter, the distance travelled without detecting any obstacles determines how far the vehicle may climb or descent when it reaches $VMP$, and also over which distance. The climb angle is defined by the slope of the $VFOV$. Recall that the maximum climb/descent height is limited as Eq.~\ref{eqn:height_climb} shows. As Fig.~\ref{droplet_ver_implement} illustrates, a vertical maneuver is only initiated if an obstacle (gray box) is detected within the Droplet region after reaching $VMP$ (middle-right and bottom-left diagram). The bottom-right diagram of this figure shows the situation where an obstacle is present in the climb region. In such cases the maximum climb/descent height is smaller than the limit specified by Eq.~\ref{eqn:height_climb}.




\subsection{Simulation results of extension to 3D}
To evaluate the advantage of implementing height control in the Droplet strategy, experiments with different pole configurations were performed. Changing the length of the poles and attaching them to either the floor or the ceiling requires the vehicle to fly over and under the poles. In the experiments the room has a height of $3$ meter and the vehicle starts its flight at $1.5$ meter altitude. Long poles extend from floor to ceiling, short poles have a length of $1.3$ meter. Short poles that are attached to the ceiling thus start at a height of $1.7$ meter. 100 flights of 600 seconds are performed for each configuration. The performance difference of the Droplet strategy with and without height control is compared in terms of total area covered. Furthermore the number of climb/descent actions is counted and the average flight altitude is computed. 

The results are listed in Table.~\ref{height_control_performance}. By replacing the long poles with short poles, the average area covered during flights increases. Note that this number does not approach 100\%. It could be further increased if specific rules would be applied to stimulate this, but the current method is not intended to do this. By applying different combinations of low poles and high poles, the average flight altitude changes accordingly. Note that the impact of the specific combination of low and high poles on the covered area is very small. Apparently the randomness of flight directions increases in this situation, leading to a higher coverage in any case.


\begin{table}[t]
\renewcommand{\arraystretch}{1.3}
\caption{Effectiveness of height control for the Droplet strategy}
\centering
\begin{tabular}{l|c c c c} 
\hline\hline
       & Cov. [\%] & \# climbs. & \#descents & altitude [m]    \\ 
\hline
5 long poles &  59.2 & 2.7 & 2.1 & 1.5 \\               
4 long 1 low poles &  64.4 & 4.1 & 2.4 & 1.7\\          
3 long 2 low poles &  70.3 & 4.8 & 3.1 & 1.8\\          
2 long 3 low poles &  78.9 & 5.9 & 3.3 & 1.9\\          
1 long 4 low poles &  85.0 & 6.0 & 2.8 & 2.0\\          
5 low poles &  86.2 & 7.3 & 3.1 & 2.0\\ 
4 low 1 high pole &  84.0 & 8.6 & 5.7 & 1.9\\           
3 low 2 high poles &  83.1 & 9.6 & 8.3 & 1.7\\          
2 low 3 high poles &  83.9 & 8.1 & 8.3 & 1.4\\ 
1 low 4 high poles &  84.5 & 7.2 & 7.4 & 1.2\\ 
5 high poles &  84.4 & 6.9 & 7.2 & 1.0 \\               

\hline\hline
\end{tabular}
\begin{tabular}{l}

\end{tabular}
\label{height_control_performance}
\end{table}






%





\ifCLASSOPTIONcaptionsoff
  \newpage
\fi



%

\bibliographystyle{IEEEtran}
\bibliography{bibliogdc}




%








\end{document}